# SOLVING NURSE SCHEDULING PROBLEM USING CONSTRAINT PROGRAMMING TECHNIQUE

Oluwaseun M. Alade[1], Akeem O. Amusat[2]

Department of Computer Science and Engineering, Faculty of Engineering and Technology, Ladoke Akintola University of Technology, Ogbomoso, Nigeria.

Corresponding Author email: oalade75@lautech.edu.ng, aoamusat@student.lautech.edu.ng

**ABSTRACT**

*Staff scheduling is a universal problem that can be encountered in many organizations, such as call centers, educational institution, industry, hospital, and any other public services. It is one of the most important aspects of workforce management strategy and the one that is most prone to errors or issues as there are many entities should be considered, such as the staff turnover, employee availability, time between rotations, unusual periods of activity, and even the last-minute shift changes. The nurse scheduling problem (NSP) is a variant of staff scheduling problems which appoints nurses to shifts as well as rooms per day taking both hard constraints, i.e., hospital requirements, and soft constraints, i.e., nurse's preferences, into account. Most algorithms used for scheduling problems fall short when it comes to the number of inputs they can handle.*

*In this paper, constraint programming was developed to solve the nurse scheduling problem. The developed constraint programming model was then implemented using python programming language.*

## 1.0 INTRODUCTION

Nurse scheduling is nothing but a weekly or monthly plan for all nurses in hospital, and is obtained by assigning shift categories to the nurses or assigning nurses to shift. Nurse Scheduling represents a task which consists of creating a schedule for the nurses in a hospital. The Nurse Scheduling Problem (NSP) is a common problem every hospital faces every day.

Constraint Programming is a relatively modern technology for solving constraint satisfaction and constraint optimization problems. It has arisen as a combination of techniques mainly coming from the operational research domain, artificial intelligence, and programming languages. The last 20 years CP has been successfully applied in different application areas, for instance to express geometric coherence in computer graphics, for the conception of complex mechanical structures, to ensure and/or restore data consistency, to locate faults in electrical engineering, and even for DNA sequencing in molecular biology (Rossi F., *Handbook of Constraint Programming*. Elsevier, 2006).

Literature on nurse rostering and scheduling is very extensive. One may refer to literature reviews on the subject that provide in-depth studies on this problem such as Burke et al (2004) and Ernst et al (2004). A wide variety of methods have been used to tackle nurse scheduling which includes: mathematical programming, constraint programming, heuristics and meta-heuristics, hybrid methods as well as simulation. Even creative methods such as auction systems have been applied to tackle nurses' preferences (De Grano et al and Medeiros et al, 2009).

Abernathy et al (1973), isolated nurse scheduling from the general staffing problem and solved it using mathematical (stochastic) programming techniques. Arthur and Ravindran (1981), propose a two-phase goal programming heuristic for the nurse scheduling problem. Darmoni et al (1995), describe a software system called "Horoplan" for scheduling nurses in a large hospital. Apart from rostering, the system also covers some short-term staffing decisions. Brusco and Jacobs (1995), combine simulated annealing and a simple local search heuristic to generate cyclical schedules for continuously operating organizations. Burke et al (1999) hybridize a tabu search approach with algorithms that are based upon human-inspired improvement techniques. Weil et al (1995), reduce the complexity of a constraint satisfaction problem by merging some constraints and by eliminating interchangeable values and thus reducing the domains.

## 2.0 METHODOLOGY

In this paper, constraint programming technique was used to solve the nurse scheduling problem. Constraint programming differ from the common primitives of other programming languages in that they do not specify a step or sequence of steps to execute but rather the properties of a solution to be found.

### 2.1 Formulation of Constraint Programming

In formulating a constraint programming problem, the constraint algorithm tries to find the best solution to the problem. If no solution or inconsistency is found, then one of the variables with domain size larger than 1 is selected and a new CSP is created for each possible assignment of this variable. The following are the steps involved in solving a constraint satisfaction problem:

- Initial variable assignment: Each time the solver makes a variable assignment, it has a couple of choices to make in order to select a solution.
- The solver evaluates the solution for optimality i.e. fitness value computation
- Backtracking: if the current solution is not optimal, the solver moves back in the search tree to try other variable assignment. This is called *backtracking.*
- Feasible solution: the solver arrives at the best feasible solution (solution with the best fitness value) when all variable has been assigned.

There's need to highlight the constraints associated with nurse scheduling problem. Constraints can be classified into two: **hard constraints** and **soft constraints.** Hard constraints are those that cannot be violated and define the feasibility of solutions. Hard constraints are concerned with the hospital's needs as opposed to the nurses' preferences. Soft constraints are desirable but not obligatory, and thus can be violated. Table 1 below shows various constraints associated with our constraint programming model:

**Table 1: Constraints and their description**

| Constraint | Description |
| --- | --- |
| HC1 | Each day is divided into three 8-hour shifts (morning, afternoon and evening). |
| HC2 | On each day, all nurses are assigned to different shifts and one nurse has the day off. |
| HC3 | Each nurse works five or six days a week. |
| HC4 | No shift is staffed by more than two different nurses in a week. |
| HC5 | If a nurse works shifts 2 or 3 on a given day, he must also work the same shift either the previous day or the following day. |

### 2.2 Mathematical Representation of the Problem

The NSP can be mathematical represented as thus:

$$\alpha \times f \left( \sum_{i=1}^{I} \sum_{k=1}^{K} \sum_{s=1}^{S} D i,k,s \right) + (1-\alpha) \times G \left( \sum_{i=1}^{I} \sum_{k=1}^{K} \sum_{s=1}^{S} D i,k,s \right)$$

where *i* represent some nurse *i*.

where *k* represents the days.

where *s* represents the shifts.

## 3. Experimental Result

The algorithm was tested with twenty nurses using four different shifts. The result for CP algorithm is shown in table 2. The algorithm was tested under four separated runs.

Table 2: this table presents a descriptive statistic on fitness values for our CP model. The statistical analysis was done using SPSS (Statistical Package for Social Sciences). From this table, it can be observed that the mean fitness value for the CP model is slightly higher than that of particle swarming optimization (PSO) algorithm. The same thing applies to the variance.

| Statistics | | CP Runtime | Fitness CP | Fitness PSO |
|---|---|---|---|---|
| N | Valid | 4 | 4 | 4 |
|   | Missing | 0 | 0 | 0 |
| Mean | | 40.7800 | 7.1975 | 6.7781 |
| Standard Error of Mean | | .04708 | .42201 | .18645 |
| Standard Deviation | | .09416 | .84401 | .47289 |
| Variance | | .009 | .712 | .530 |

Table 4.1: Shifts generated by constraint programming model for 10 nurses.

| Nurse/Day | 1 | 2 | 3 | 4 | 5 | 6 | 7 | 8 | 9 | 10 | 11 | 12 | 13 | 14 | 15 | 16 | 17 | 18 | 19 | 20 | 21 | 22 | 23 | 24 | 25 | 26 | 27 | 28 | 29 |
|---|---|---|---|---|---|---|---|---|---|---|---|---|---|---|---|---|---|---|---|---|---|---|---|---|---|---|---|---|---|
| N1 | A | N | O | O | N | A | M | M | N | N | N | M | M | M | N | O | O | M | O | A | O | A | N | N | N | N | N | M | A |
| N2 | O | O | M | N | M | A | N | A | N | N | A | N | M | A | O | A | M | A | O | M | A | A | O | O | N | N | O | A | O |
| N3 | N | M | O | A | N | N | O | O | O | N | A | A | M | O | A | O | O | O | O | M | N | A | A | M | O | A | A | A | M |
| N4 | N | N | N | N | O | A | N | M | A | N | A | A | A | O | N | A | O | N | M | O | A | N | A | A | A | N | O | O | O |
| N5 | A | A | N | A | A | M | A | M | A | A | A | N | O | O | M | A | M | A | M | A | M | N | O | O | A | A | M | N |   |
| N6 | M | O | A | M | N | A | M | O | O | M | A | M | A | A | O | A | M | N | A | A | N | A | N | A | N | N | O | N | M |
| N7 | M | O | N | N | M | O | N | A | A | M | O | A | O | M | O | O | M | N | O | O | N | O | M | A | M | O | N | O | O |
| N8 | A | M | A | N | O | A | M | M | O | O | M | M | A | M | O | N | N | O | A | M | A | N | O | N | A | O | O | M | O |
| N9 | O | O | O | N | O | O | N | A | M | A | O | A | N | N | A | A | A | M | A | N | O | M | O | N | O | M | O | A | N |
| N10 | N | O | N | O | A | N | A | N | N | N | O | M | M | A | O | N | M | N | M | O | A | A | A | O | N | O | N | O | A |

## 4.1 CONCLUSION

In this project report, we have made use of constraint programming (CP) technique to solve nurse scheduling problem (NSP). The aim of this problem is to maximize the fairness of the schedule, while respecting all hard constraints. In regards with the result obtained after various test on different datasets, the CP technique shows its ability in finding optimal solution to NSP with higher footprint on computational resources. The CP technique can be summarily described below as:

- CP is a general technique, can encapsulate a lot of work.
- CP allows the use of symbolic representation.
- The performance of search depends on the number input.

## 4.2 FURTHER WORK

After reviewing the performance of CP technique on NSP, the following future studies can be done:

- Add more constraints to the model.
- Implement the CP model on GPU (graphics processing unit) to minimize the footprint on system resources and also provide GUI (graphical user interface) for easy use by naive users.